\documentclass[11pt]{article}

\usepackage[final]{acl}

\usepackage{times}
\usepackage{latexsym}
\usepackage[T1]{fontenc}
\usepackage[utf8]{inputenc}
\usepackage{microtype}
\usepackage{inconsolata}
\usepackage{graphicx}
\usepackage{tikz}
\usepackage{multirow}
\usepackage[table]{xcolor}
\usepackage{booktabs}
\usepackage{amsmath}
\usepackage{enumitem}
\usepackage{comment}
\usepackage{siunitx}
\usepackage{fontawesome5}
\usetikzlibrary{arrows.meta}

\title{Reader Proficiency Shapes Layer-wise Surprisal Profiles}

\author{Akio Hayakawa \quad \quad \quad Horacio Saggion\\
  \begin{tabular}{c}
      Universitat Pompeu Fabra\\
      Barcelona, Spain\\
      \texttt{\{akio.hayakawa,  horacio.saggion\}@upf.edu}
  \end{tabular}
}

\begin{document}
\maketitle

\begin{abstract}

Reading behaviour varies not only with linguistic input, but also with reader proficiency. In this study, we investigate whether the layer-wise relationship between surprisal from large language models (LLMs) and human gaze behaviour differs across readers with different levels of proficiency and across gaze measures.
Using eye-tracking data from the MECO L2 corpus, we compare readers with high and low vocabulary proficiency on first-pass gaze duration (FPGD) and total gaze duration (TGD). We quantify the distribution of the predictive power of surprisal across model layers using Predictive Depth.
Across 12 tested LLMs, we find that readers with lower vocabulary proficiency tend to show deeper Predictive Depth for FPGD, while this difference is smaller for TGD. Also, TGD itself shows deeper Predictive Depth than FPGD in both proficiency groups. These patterns suggest that where predictive power is concentrated across LLM layers may be related to the timing and breadth of the reading processes captured by different gaze measures, and that this relationship can vary with reader proficiency.
Our leave-one-out analysis further shows that the advantage of informative internal layers extends to unseen texts, although the practical improvements in prediction are limited. Overall, our results show that layer-wise LLM surprisal provides a useful perspective on variation in reading behaviour across both reader groups and gaze measures.\footnote{Code will be available at \url{https://github.com/ahaya3776/proficiency_surprisal}.}

\end{abstract}

\section{Introduction}

While text difficulty is often treated as a fixed property of the text, understanding the reading process requires considering the interaction between readers and linguistic structures. This perspective is particularly important in fields such as text simplification, which aims to enhance reader comprehension by rewriting complex texts into more accessible forms \citep{siddharthan2014survey, alva-manchego-etal-2020-data}.
However, traditional methods for evaluating text difficulty often rely on surface-level metrics, such as Flesch-Kincaid Grading Levels \citep{kincaid1975derivation}, or comprehension accuracy \citep{agrawal-carpuat-2024-text}. While these measures can indicate if a text is linguistically simple or if a reader eventually understands it, they provide limited information about how the text is processed during reading. This is of paramount importance because readers with different levels of proficiency may process the same text differently.

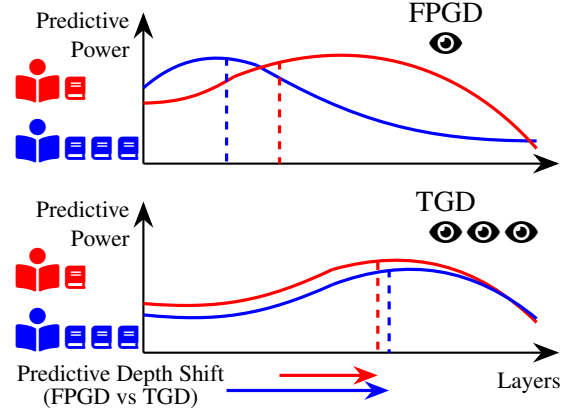
\begin{figure}[t]
\centering
\begin{tikzpicture}
    \draw[-{Stealth[scale=1.5]}, black, thick] (3, 0.5) -- (8.5, 0.5);
    \draw[-{Stealth[scale=1.5]}, black, thick] (3, 0.5) -- (3, 2.5);
    \node[align=right, scale=0.8] at (2.2, 2.2) {Predictive\\Power};
    \node[scale=1.5, red] at (1.6, 1.6) {\faBookReader};
    \node[scale=0.8, red] at (2.1, 1.5) {\faBook};
    \node[scale=1.5, blue] at (1.6, 0.8) {\faBookReader};
    \node[scale=0.8, blue] at (2.1, 0.7) {\faBook};
    \node[scale=0.8, blue] at (2.45, 0.7) {\faBook};
    \node[scale=0.8, blue] at (2.8, 0.7) {\faBook};
    \node[scale=1.0] at (7.0, 2.5) {FPGD};
    \node[scale=1.0] at (7.0, 2.1) {\faEye};

    \draw[very thick, blue] (3, 1.5) to[bend left=30] (4.5, 1.8);
    \draw[very thick, blue] (4.5, 1.8) to[bend right=15] (8.2, 0.8);
    \draw[very thick, red] (3, 1.3) to[bend right=15] (4.2, 1.65);
    \draw[very thick, red] (4.2, 1.65) to[bend left=35] (8.2, 0.7);
    \draw[very thick, dashed, blue] (4.10, 1.9) to (4.10, 0.5);
    \draw[very thick, dashed, red] (4.8, 1.85) to (4.8, 0.5);

    \draw[-{Stealth[scale=1.5]}, black, thick] (3, -2) -- (8.5, -2);
    \node[align=center, scale=0.8] at (8.1, -2.4) {Layers};
    \draw[-{Stealth[scale=1.5]}, black, thick] (3, -2) -- (3, 0);
    \node[align=right, scale=0.8] at (2.2, -0.3) {Predictive\\Power};
    \node[scale=1.5, red] at (1.6, -0.9) {\faBookReader};
    \node[scale=0.8, red] at (2.1, -1.0) {\faBook};
    \node[scale=1.5, blue] at (1.6, -1.7) {\faBookReader};
    \node[scale=0.8, blue] at (2.1, -1.8) {\faBook};
    \node[scale=0.8, blue] at (2.45, -1.8) {\faBook};
    \node[scale=0.8, blue] at (2.8, -1.8) {\faBook};
    \node[scale=1.0] at (7.0, 0.0) {TGD};
    \node[scale=1.0] at (7.0, -0.4) {\faEye};
    \node[scale=1.0] at (7.5, -0.4) {\faEye};
    \node[scale=1.0] at (8.0, -0.4) {\faEye};

    \draw[very thick, red] (3, -1.35) to[bend right=15] (5.5, -0.90);
    \draw[very thick, red] (5.5, -0.90) to[bend left=30] (8.2, -1.60);
    \draw[very thick, blue] (3, -1.50) to[bend right=15] (5.5, -1.1);
    \draw[very thick, blue] (5.5, -1.1) to[bend left=30] (8.2, -1.55);
    \draw[very thick, dashed, red] (6.10, -0.8) to (6.10, -2.0);
    \draw[very thick, dashed, blue] (6.25, -0.90) to (6.25, -2.0);

    \draw[-{Stealth[scale=1.1]}, red, very thick] (4.80, -2.3) -- (6.10, -2.3);
    \draw[-{Stealth[scale=1.1]}, blue, very thick] (4.10, -2.5) -- (6.25, -2.5);
    \node[align=center, scale=0.8] at (2.7, -2.3) {Predictive Depth Shift};
    \node[align=center, scale=0.8] at (2.7, -2.6) {(FPGD vs TGD)};

\end{tikzpicture}
\caption{Illustration of the distribution of predictive power along LLM layers. Readers with limited vocabulary knowledge (red) tend to show deeper Predictive Depth for FPGD compared to readers with higher vocabulary knowledge (blue). The difference in Predictive Depth is smaller for TGD, which itself shows deeper Predictive Depth than FPGD in both reader groups.}
\end{figure}

In this context, surprisal from large language models (LLMs) provides a useful way to examine human reading behaviour. While surprisal has been widely utilized to predict human reading times \citep{smith2013effect,kuribayashi-etal-2022-context, shain2024large}, recent findings suggest that different internal layers of LLMs have different predictive power for human reading times \citep{kuribayashi-etal-2025-large}.
Importantly, different reading measures are best predicted by surprisal from different parts of LLMs.
For example, measures which capture relatively fast responses, such as gaze duration, tend to be better predicted by surprisal from shallower layers, whereas measures which capture slower responses show stronger predictive power in deeper layers.
Among eye-tracking measures, first-pass gaze duration (FPGD) and total gaze duration (TGD) capture different spans of reading behaviour. FPGD reflects the initial pass over a word, while TGD includes later and repeated fixations.
This pattern may be related to differences in the timing and breadth of cognitive processing captured by each measure.

While previous studies have mainly focused on native English speakers, the potential of this approach becomes evident when considering the diversity of reader proficiency. Cognitive models of reading suggest that less-proficient readers may engage in compensatory processing \citep{stanovich-1980-toward} when faced with linguistic obstacles, such as difficult vocabulary or syntax. In such cases, they may rely more on contextual information to support comprehension. This raises the possibility that reader proficiency may also be an important factor in determining which internal layers best explain reading behaviour.

In this study, we examine how the layer-wise relationship between LLM surprisal and reading behaviour varies with reader proficiency and gaze measure, focusing on FPGD and TGD. To summarise the layer-wise distribution of predictive power, we introduce Predictive Depth, which measures how the predictive power of surprisal is distributed across LLM layers.

We find that the layer-wise relationship between LLM surprisal and reading behaviour varies both with reader proficiency and gaze measure. For FPGD, readers with lower vocabulary proficiency tend to show deeper Predictive Depth, while this difference is substantially reduced for TGD. Also, TGD generally shows deeper Predictive Depth than FPGD across both proficiency groups. A leave-one-out analysis further shows that the predictive advantage of selected internal layers extends to unseen texts, although the improvements are small.

Our main contributions are as follows:
\begin{itemize}
    \item We show that the layer depth at which surprisal best predicts reading behaviour varies with vocabulary proficiency.
    \item We show that this relationship also differs between FPGD and TGD, revealing a clear effect of gaze measure.
    \item We show that surprisal provides weak to modest predictive gains on unseen texts, with the selected internal layer generally outperforming the final layer.
\end{itemize}

\section{Related Work}
\label{sec:theory}

\subsection{Cognitive Models of Reading}

Reading involves the coordination of lower-level lexical access with higher-level semantic integration \citep{kintsch1988role}. A fundamental distinction between proficient and less-proficient readers lies in the automaticity of these processes. Proficient readers can access lexical information with relatively little conscious effort \citep{laberge1974toward}. This automaticity leaves more cognitive resources available for higher-level semantic integration \citep{perfetti2002lexical}. In contrast, when lexical access is less automatic, readers may rely more on contextual information to support word recognition and comprehension \citep{stanovich-1980-toward}.
While contextual information can support comprehension, greater reliance on context may also require readers to integrate information across a broader context \citep{harrington1992l2, CARRETTI2009246}.
Therefore, differences in proficiency can affect not only whether a text is understood, but also how linguistic information is processed during reading.

These differences are also reflected in eye-movement behaviour. Measures such as FPGD and TGD provide observable traces of the underlying cognitive processes involved in reading \citep{rayner1998eye}. Examining variation in these measures can help us investigate how processes such as lexical access and semantic integration differ across readers with different levels of proficiency.

\subsection{Layer-wise Surprisal and Reading Measures}
In computational linguistics, surprisal is defined as the negative log-probability of a word given its preceding context \citep{hale-2001-probabilistic, levy2008expectation}. Surprisal from language models has long been used to model human reading behaviour \citep{smith2013effect, kuribayashi-etal-2022-context, shain2024large}. In recent cognitive modeling frameworks, the unique predictive power of surprisal is typically quantified by the increase in log-likelihood ($\Delta LL$) when surprisal is added to a regression model containing baseline linguistic features \citep{oh2022comparison}. More recently, studies have shown that surprisal from the final layer of modern large language models (LLMs) does not always provide the best explanation for human reading times \citep{oh-schuler-2023-surprisal}, and that surprisal from internal layers can improve predictive performance \citep{kuribayashi-etal-2025-large}.

This has motivated closer examination of how predictive power varies across layer depth. \citet{kuribayashi-etal-2025-large} examined several behavioural measures, including gaze durations, self-paced reading times, N400 responses, and MAZE processing times. They found that relatively fast measures such as gaze duration were better predicted by surprisal from shallower layers, whereas slower measures such as N400 and MAZE showed stronger predictive power in deeper layers. This suggests that the depth at which LLM surprisal is most predictive may vary with the temporal or processing scope of the human measure.
Related probing studies have also shown that deeper layer representations tend to encode more contextualized information \citep{peters-etal-2018-dissecting, vulic-etal-2020-probing, jin-etal-2025-exploring}. These findings are consistent with the view that layer depth may reflect differences in the amount or scope of contextual information encoded across layers.

Recent studies further suggest that this relationship depends on the type of processing being examined. \citet{kuribayashi-etal-2026-dual} found that shallower layers better capture naturalistic reading, whereas deeper layers better capture processing difficulty in syntactically challenging constructions.
\citet{tsipidi-etal-2026-probing} similarly found that the strongest predictors vary across eye-tracking measures, with representations from shallower layers performing particularly well for fast response measures and different patterns occurring for slower measures. These findings motivate examining how the predictive power of surprisal changes across layer depth and across gaze measures.

\subsection{Research Questions}
The literature reviewed above suggests two relevant sources of variation in reading behaviour. First, readers with different levels of proficiency may differ in their reliance on lexical and contextual information during reading. Second, the predictive power of LLM surprisal varies across layer depth and across measures that capture different aspects of the reading process. We therefore examine whether the layer-wise distribution of predictive power varies across reader proficiency and across gaze measures. We address this question through the following research questions:
\begin{itemize}
    \setlength{\itemsep}{1pt}
    \setlength{\parskip}{0pt}
    \item \textbf{RQ1}: Does reader proficiency affect the distribution of predictive power of LLM surprisal across layer depth?
    \item \textbf{RQ2}: Does this distribution differ between first-pass gaze duration and total gaze duration?
\end{itemize}

\section{Method}

We investigate the relationship between human gaze patterns and LLM internal representations through a layer-wise surprisal framework. We focus on non-native English readers for the experiment.

\subsection{Dataset}
We utilize the \textbf{MECO L2} dataset (Waves 1 and 2) \citep{kuperman2023text, kuperman2025new}, which provides eye-tracking data from L2 English speakers with 21 different native language backgrounds who read the same English texts. The participants read 12 encyclopedic texts, ranging from 100 to 200 words in length, and answered comprehension questions after each text.

Our analysis uses two word-level gaze measures, FPGD and TGD. FPGD is defined as the total fixation time on a word during the first pass before eyes move away from that word, capturing relatively early processing during reading. TGD is defined as the total fixation time on a word, including all subsequent fixations, capturing gaze behaviour over a broader processing span. FPGD and TGD are computed over the same eye-tracking observations\footnote{We exclude the first word of each sentence to avoid initial processing noise.}. Both measures are log-transformed prior to statistical analysis, and log-FPGD and log-TGD are used as dependent variables in our main analyses.

\subsection{Reader Profiling}
To investigate the impact of vocabulary proficiency, we use LexTALE scores \citep{lemhofer2012introducing}, which are included in the MECO L2 metadata. LexTALE is a lexical decision task in which participants identify whether a given string is a real word or not, providing an efficient estimate of vocabulary size. Based on these scores, we categorize readers into quartiles and focus our analysis on the top (High-Lex) and bottom (Low-Lex) groups.

\begin{table}[t]
    \centering
    \footnotesize
    \begin{tabular}{cccc}
         & All & High-Lex & Low-Lex \\
        \midrule
        LexTALE Score & - & $\ge$ 83.75 & $<$ 66.25 \\
        Group Size & 1001 & 276 & 228 \\
        Total Trials & 8597 & 2519 & 1845 \\
        \bottomrule
    \end{tabular}
    \caption{Statistics of reader groups from MECO L2. Trials are the number of times each participant read each text.}
    \label{tab:stats}
\end{table}

\autoref{tab:stats} summarizes the participants in the High-Lex and Low-Lex groups. These groups roughly correspond to advanced (C1-C2) and intermediate (B1-lower B2) proficiency levels, respectively \citep{lemhofer2012introducing}.

\subsection{LLMs and Layer-wise Surprisal Calculation}
We utilize the internal representations of several LLMs, including Gemma 3 (4B, 12B, 27B) \citep{gemmateam2025gemma3technicalreport}, Qwen 2.5 (7B, 14B, 32B, 72B) \citep{qwen2025qwen25technicalreport}, OLMo 2 (7B, 13B, 32B) \citep{olmo20252olmo2furious}, and Llama 3.1 (8B, 70B) \citep{grattafiori2024llama3herdmodels}. We use base models instead of instruction-tuned models.

Building on \citet{kuribayashi-etal-2025-large}, we apply the Logit-Lens technique\footnote{\url{https://www.lesswrong.com/posts/interpreting-gpt-the-logit-lens}, visited on August 30, 2026.} to extract layer-wise information. This method applies the LLM's final output head directly to the hidden states at each layer. This allows us to derive a probability distribution for each word and calculate its surprisal at every layer, rather than only at the final layer. We calculate these values by providing the full context of each text to the LLMs. For words split into multiple subwords, individual surprisal values are summed to represent the surprisal of the entire word.

Through this process, we obtain a surprisal value for every word at every layer of each LLM. These values serve as the basis for evaluating layer-wise predictive power in the following analysis.

\subsection{Statistical Analysis}
\subsubsection{LMM Analysis}
We use linear mixed-effects models (LMMs) \citep{BAAYEN2008390}, a type of regression model, to quantify the layer-wise relationship between LLM surprisal and gaze duration while accounting for repeated observations from participants and reading trials. Here, the dependent variable is log-FPGD or log-TGD, and the explanatory variables include word length, word frequency, and surprisal from LLM layers.
We conduct the layer-wise analyses separately for the High-Lex and Low-Lex reader groups.

For each reader group $G$ and layer $k$, we compare a baseline LMM with a full LMM that additionally includes current-word surprisal. Let $Y$ denote the log-transformed gaze measure, either log-FPGD or log-TGD. The baseline LMM is specified as:
\begin{multline*}
M^{Base}_{G,k}: Y \sim L_{w_i} + L_{w_{i-1}} + L_{w_{i-2}} \\
+ F_{w_i} + F_{w_{i-1}} + F_{w_{i-2}} + S_{w_{i-1}, k} + S_{w_{i-2}, k} \\
+ \mathrm{Random\ Effects},
\end{multline*}
where $w_i$ denotes the current word, and $w_{i-1}$ and $w_{i-2}$ denote the two preceding words. $L$ and $F$ denote the word length (the number of characters in a word) and Zipf frequency, respectively. Frequency estimates are obtained using the \texttt{wordfreq} library \citep{robyn_speer_2022_7199437}. $S_{w_{i-1}, k}$ and $S_{w_{i-2}, k}$ denote the surprisal values of the two preceding words extracted from layer $k$. Random effects include random intercepts for participants and reading trials.

The full LMM adds the surprisal of the current word from the same layer $k$ as the ninth variable:
$$
M^{Full}_{G,k}: M^{Base}_{G,k} + S_{w_i, k}.
$$

Including information from the preceding words in the baseline controls for spillover effects \citep{rayner1998eye} and allows us to isolate the unique contribution of current-word surprisal at each layer. The same random-effects structure is used for the baseline and full LMMs.

Since the baseline and full LMMs differ in their fixed-effects structure, we fit all LMMs using maximum-likelihood (ML) instead of restricted maximum-likelihood (REML) estimation for comparison. For each reader group $G$ and layer $k$, we calculate the unique contribution of current-word surprisal as the increase in LMM log-likelihood:
$$
\Delta LL_{G,k} = LL(M^{Full}_{G,k}) - LL(M^{Base}_{G,k}).
$$
Here, $LL(M)$ is the log-likelihood of the regression model $M$, which measures how well the regression model fits the observed data.
Larger values of $\Delta LL_{G,k}$ indicate a greater contribution of current-word surprisal to explaining the corresponding gaze measure at that layer. Since the full LMM is a nested version of the baseline LMM, the maximized $\Delta LL_{G,k}$ is theoretically non-negative\footnote{In practice, we observe some small negative values due to numerical optimization, which we treat as zero in our analysis.}. LMMs are fitted using the \texttt{MixedLM} class from the \texttt{statsmodels} library \citep{seabold2010statsmodels}.

\begin{figure*}[t]
    \centering
    \includegraphics[width=\linewidth]{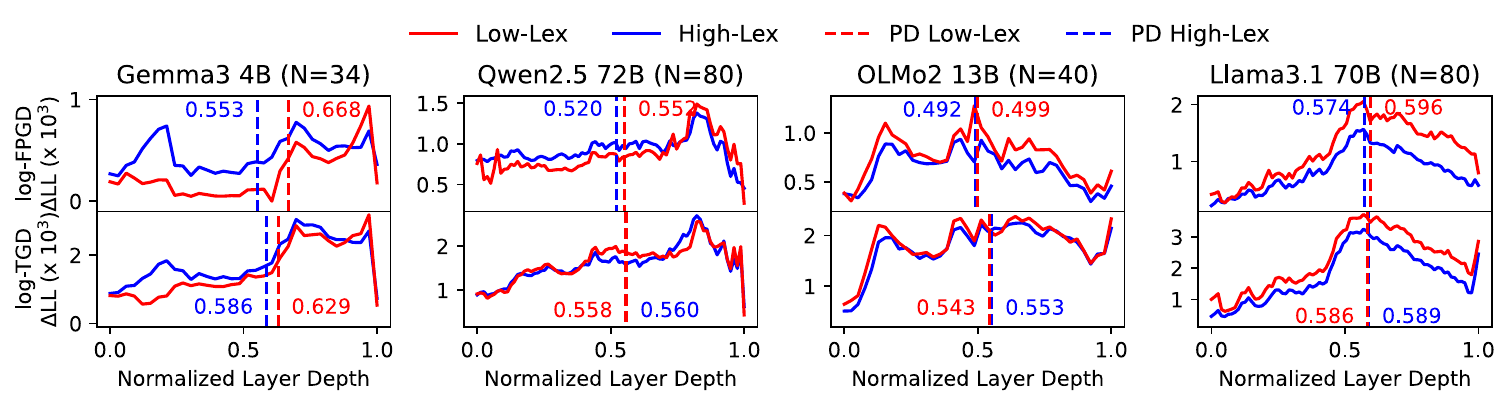}
    \caption{Plots of layer-wise $\Delta LL$ for both FPGD (top) and TGD (bottom) from each LLM family. $\Delta LL$ values on the y-axis are normalized per token and multiplied by 1000 for better visualization.}
    \label{fig:deltall}
\end{figure*}

\subsubsection{Predictive Depth}
To summarize where the predictive power of surprisal is distributed across layers, we define Predictive Depth $PD$ as the normalized weighted average of layers:
$$
PD_G = \frac{\sum\limits_{k=1}^N k \cdot \Delta LL_{G,k}}{N \cdot \sum\limits_{k=1}^N \Delta LL_{G,k}},
$$
where the weights are $\Delta LL_{G,k}$ and $N$ is the total number of layers.

Predictive Depth represents the center of the layer-wise distribution of the predictive power of current-word surprisal to explain the corresponding gaze measure. Larger Predictive Depth indicates that the predictive power is distributed towards relatively deeper layers. We compute Predictive Depth separately for log-FPGD and log-TGD for comparing both across reader groups and across gaze measures.

\setlength{\tabcolsep}{4pt}
\begin{table*}[t]
    \centering
    \begin{tabular}{rr|r|S[table-format=.3, retain-explicit-plus, mode=text]S[table-format=+1.3, retain-explicit-plus, mode=text]S[table-format=+1.3, retain-explicit-plus, mode=text]|S[table-format=+1.3, retain-explicit-plus, mode=text]S[table-format=+1.3, retain-explicit-plus, mode=text]S[table-format=+1.3, retain-explicit-plus, mode=text]|S[table-format=+1.3, retain-explicit-plus, mode=text]S[table-format=+1.3, retain-explicit-plus, mode=text]}
        & & & \multicolumn{3}{c|}{FPGD} & \multicolumn{3}{c|}{TGD} & \multicolumn{2}{c}{TGD-FPGD} \\
        {LLM} & {Size} & {N} & {Low} & {High} & {L-H} & {Low} & {High} & {L-H} & {Low} & {High} \\
        \midrule
        \multirow{3}{*}{Gemma 3} & 4B & 34 & 0.668 & 0.553 & \textbf{+0.115} & 0.629 & 0.586 & \textbf{+0.043} & -0.039 & \textbf{+0.033} \\
         & 12B & 48 & 0.613 & 0.548 & \textbf{+0.065} & 0.678 & 0.599 & \textbf{+0.079} & \textbf{+0.065} & \textbf{+0.051} \\
         & 27B & 62 & 0.523 & 0.549 & -0.026 & 0.674 & 0.673 & \textbf{+0.001} & \textbf{+0.151} & \textbf{+0.124} \\
         \midrule
        \multirow{4}{*}{Qwen 2.5} & 7B & 28 & 0.558 & 0.549 & \textbf{+0.009} & 0.565 & 0.587 & -0.022 & \textbf{+0.000} & \textbf{+0.039} \\
         & 14B & 48 & 0.520 & 0.507 & \textbf{+0.014} & 0.558 & 0.559 & -0.001 & \textbf{+0.038} & \textbf{+0.052} \\
         & 32B & 64 & 0.509 & 0.485 & \textbf{+0.024} & 0.545 & 0.540 & \textbf{+0.004} & \textbf{+0.036} & \textbf{+0.055} \\
         & 72B & 80 & 0.552 & 0.520 & \textbf{+0.031} & 0.558 & 0.560 & -0.003 & \textbf{+0.006} & \textbf{+0.040} \\
         \midrule
        \multirow{3}{*}{OLMo 2} & 7B & 32 & 0.525 & 0.520 & \textbf{+0.005} & 0.545 & 0.560 & -0.015 & \textbf{+0.020} & \textbf{+0.040} \\
         & 13B & 40 & 0.499 & 0.492 & \textbf{+0.008} & 0.543 & 0.553 & -0.010 & \textbf{+0.044} & \textbf{+0.061} \\
         & 32B & 64 & 0.534 & 0.535 & -0.001 & 0.561 & 0.575 & -0.013 & \textbf{+0.027} & \textbf{+0.040} \\
         \midrule
        \multirow{2}{*}{Llama 3.1} & 8B & 32 & 0.640 & 0.620 & \textbf{+0.020} & 0.658 & 0.676 & -0.018 & \textbf{+0.018} & \textbf{+0.056} \\
         & 70B & 80 & 0.596 & 0.574 & \textbf{+0.021} & 0.586 & 0.589 & -0.003 & -0.010 & \textbf{+0.015} \\
        \bottomrule
    \end{tabular}
    \caption{Predictive Depth for FPGD and TGD in the Low-Lex and High-Lex groups across all tested LLMs. $N$ represents the number of layers in each model. The last two columns show the difference in $PD$ between TGD and FPGD for each group. Positive differences are shown in bold.}
    \label{tab:pd}
\end{table*}

\section{Results}

\subsection{Layer-wise Predictive Profiles}
Before comparing Predictive Depth across reader groups and gaze measures, we first examine the layer-wise predictive power of surprisal. As a result, surprisal from internal layers generally provides better explanations for gaze durations than surprisal from the final output layer, especially for log-FPGD, while the predictive power is not uniformly distributed across layers.

\autoref{fig:deltall} shows representative $\Delta LL$ curves from each tested LLM family for both FPGD and TGD, while the full set of plots can be found in {\renewcommand{\sectionautorefname}{Appendix}\autoref{sec:plot_surprisal}}.
Gemma 3 4B and OLMo 2 13B exhibit multiple peaks across layers, while Qwen 2.5 72B and Llama 3.1 70B show stronger predictive power towards deeper layers. The detailed profiles also differ between FPGD and TGD, with some LLMs showing stronger predictive power near the final layers for TGD. Despite these broad tendencies, the detailed profiles vary across LLMs and gaze measures, indicating that gaze-predictive surprisal is distributed differently across LLM architectures.

\subsection{Impact of Vocabulary Proficiency}
Vocabulary proficiency is systematically associated with the Predictive Depth of surprisal for FPGD. As summarized in \autoref{tab:pd}, $PD_{\mathrm{Low}}$ is deeper than $PD_{\mathrm{High}}$ for 10 out of 12 tested models for FPGD, with a mean difference of +0.024. Although the magnitude of the difference varies across LLMs, the overall pattern is observed across multiple LLM families. As illustrated in \autoref{fig:deltall} (top), Low-Lex readers often show relatively stronger predictive power in deeper layers, shifting Predictive Depth towards greater depths.

This difference between proficiency groups is substantially weaker for TGD. $PD_{\mathrm{Low}}$ is deeper than $PD_{\mathrm{High}}$ in only 4 out of 12 LLMs, and the mean difference is +0.003. Therefore, while readers with less vocabulary tend to show deeper Predictive Depth for FPGD, the corresponding difference is small for TGD.

\subsection{FPGD vs TGD}
We also compare the Predictive Depth between FPGD and TGD within each reader group. As summarized in \autoref{tab:pd}, TGD has a deeper Predictive Depth than FPGD for all 12 LLMs in High-Lex, with a mean difference of +0.050. The same tendency is observed in Low-Lex, where TGD has a deeper Predictive Depth than FPGD for 10 out of 12 LLMs, with a mean difference of +0.030. Therefore, the shift towards deeper layers is observed in both groups, but its magnitude is larger in High-Lex, consistent with the smaller proficiency difference observed in TGD.

\section{Generalization to Unseen Texts}
The preceding analyses show that the layer-wise relationship between LLM surprisal and reading behaviour varies with both vocabulary proficiency and gaze measure. An important question is whether these relationships generalize beyond the texts used for LMM fitting. To address this question, we conduct a leave-one-out (LOO) analysis. Since MECO L2 contains 12 texts, we fit the LMMs on 11 texts and evaluate the learned relationships on the remaining unseen text in each fold.

\subsection{Leave-One-Out Evaluation}
For this LOO analysis, we focus on FPGD, our primary gaze measure, to evaluate whether the main finding on vocabulary proficiency transfers to unseen texts.
For each LOO fold, LMMs for High-Lex and Low-Lex readers are fitted and evaluated separately. The internal layer used for evaluation is selected exclusively from 11 training texts, based on the layer-wise $\Delta LL$ obtained from LMMs. Then, we evaluate the selected internal layer on the held-out text, and compare its predictive performance with that of the final LLM layer and a surface-feature baseline.

Specifically, we compare the following three LMMs:
\begin{multline*}
M^{\mathrm{Surface}}: \log\mathrm{FPGD} \sim L_{w_i} + L_{w_{i-1}} + L_{w_{i-2}} \\
+ F_{w_i} + F_{w_{i-1}} + F_{w_{i-2}} + \mathrm{Random\ Effects},
\end{multline*}
$$
M^{\mathrm{Final}} = M^{\mathrm{Surface}} + S_{w_i, N} + S_{w_{i-1}, N} + S_{w_{i-2}, N},
$$
$$
M^{\mathrm{Sel}} = M^{\mathrm{Surface}} + S_{w_i, k^{*}} + S_{w_{i-1}, k^{*}} + S_{w_{i-2}, k^{*}},
$$
where $M^{\mathrm{Final}}$ denotes the LMM using the final LLM layer $N$, and $M^{Sel}$ denotes the LMM using the selected internal layer $k^{*}$ with the largest $\Delta LL$ on the training texts.

This predictive analysis slightly differs from the main layer-profile analysis. In the main analysis, previous-word surprisal is included in the baseline LMM so that $\Delta LL$ isolates the unique contribution of current-word surprisal. In the LOO evaluation, surprisal of current and previous words from the selected layer is used jointly to predict log-FPGD, since the goal of this analysis is to evaluate how well the layer representations predict FPGD on an unseen text.

In this LOO analysis, while LMM fitting on the 11 texts is performed on individual-level observations, evaluation on the held-out text is based on word-level averages within each proficiency group with fixed effects only. Therefore, the resulting scores reflect word-level variation at the group level rather than individual-level differences.

\subsection{Evaluation Metrics}
We evaluate predictive performance on the held-out text in two ways. First, we calculate mean absolute error (MAE) between the observed and predicted log-FPGD for each word. Second, we compute Spearman correlations between residual word-level log-FPGD and current-word surprisal from the selected internal layer and the final layer.

MAE measures how closely each LMM predicts the observed gaze durations for words in the held-out text. We calculate MAE for each LOO fold and then average the MAE values across the 12 folds.
The correlation analysis evaluates whether layer-specific surprisal is associated with word-level differences that are not captured by the Surface LMM. To obtain these differences, we define residual log-FPGD $r_{G, w}$ as:
$$
r_{G, w} = \overline{Y}_{G, w} - \hat{Y}^{\mathrm{Surface}}_{G, w},
$$
where $\overline{Y}_{G, w}$ is the observed log-FPGD averaged within each group $G$ for word $w$, and $\hat{Y}^{\mathrm{Surface}}_{g, w}$ is the predicted log-FPGD from the Surface LMM. We then compute Spearman correlations between $r_{G, w}$ and current-word surprisal separately for the selected internal layer and the final layer. We then report the mean correlation coefficients across the 12 folds.

\subsection{Generalization Results}
\autoref{tab:loto} summarizes the LOO evaluation results.
Across LLMs, adding surprisal generally produces lower MAE than the Surface LMM. The Selected LMM achieves lower MAE than the Surface LMM for all 12 LLMs in both proficiency groups, and also outperforms the Final LMM in 21 out of 24 cases. However, the absolute reductions in MAE are small across LLMs and proficiency groups. Thus, although surprisal from the selected internal layer generally performs better on unseen texts, the overall improvement in MAE remains small.

A similar pattern is observed in the correlation analysis.
Residual log-FPGD shows stronger correlations with surprisal from the selected internal layer than with surprisal from the final layer in 19 out of 24 cases across LLMs and proficiency groups. Nevertheless, the absolute correlation values remain weak to modest, with the strongest condition reaching approximately $\rho =0.24$, indicating that layer-specific surprisal captures only part of the word-level variation that remains after the Surface LMM. Together with the MAE results, this suggests that the layer-wise relationships identified on the training texts transfer to unseen texts, but that the strength of this transfer is limited.

\setlength{\tabcolsep}{2pt}
\begin{table}[t]
    \centering
    \fontsize{10}{11}\selectfont
    \begin{tabular}{cc|cc|S[table-format=-1.3, table-alignment-mode=format,detect-weight=true, mode=text]c}
        \toprule
        \multicolumn{3}{c}{\textit{Low-Lex}} & \multicolumn{3}{c}{} \\
        \midrule
        & & \multicolumn{2}{c|}{MAE} & & \\
        & & \multicolumn{2}{c|}{$(\times 100)$} & \multicolumn{2}{c}{$\rho$}\\
        \multicolumn{2}{c|}{LLM}  & Final & Selected & Final & Selected\\
        \midrule
        \multicolumn{2}{c|}{(Surface)} & \multicolumn{2}{c|}{8.96} & \multicolumn{2}{c}{-} \\
        \midrule
        \multirow{3}{*}{Gemma 3} & 4B & 8.94 & 8.87 & 0.089 & 0.117\\
         & 12B & 8.95 & 8.87 & -0.030 & 0.125\\
         & 27B & 8.95 & 8.88 & -0.046 & 0.161\\
        \midrule
        \multirow{4}{*}{Qwen 2.5} & 7B & 9.00 & 8.83 & 0.064 & 0.140\\
         & 14B & 8.94 & 8.93 & -0.029 & 0.095\\
         & 32B & 8.90 & 8.89 & \bfseries 0.149 & 0.161\\
         & 72B & 8.99 & 8.80 & 0.073 & 0.142\\
        \midrule
         \multirow{3}{*}{OLMo 2} & 7B & 8.92 & 8.84 & 0.100 & 0.170\\
         & 13B & 8.92 & 8.95 & 0.118 & 0.116\\
         & 32B & 8.92 & 8.91 & 0.132 & 0.124\\
        \midrule
         \multirow{2}{*}{Llama 3.1} & 8B & 8.90 & 8.86 & 0.098 & 0.168\\
         & 70B & \textbf{8.89} & \textbf{8.67} & 0.135 & 0.239\\
        \bottomrule

        \toprule
        \multicolumn{3}{c}{\textit{High-Lex}} & \multicolumn{3}{c}{} \\
        \midrule
        \multicolumn{2}{c|}{LLM}  & Final & Selected & Final & Selected\\
        \midrule
        \multicolumn{2}{c|}{(Surface)} & \multicolumn{2}{c|}{8.31} & \multicolumn{2}{c}{-} \\
        \midrule
        \multirow{3}{*}{Gemma 3} & 4B & 8.29 & 8.25 & -0.012 & 0.132\\
         & 12B & 8.29 & 8.28 & -0.027 & 0.113\\
         & 27B & 8.30 & 8.27 & -0.044 & 0.125\\
         \midrule
        \multirow{4}{*}{Qwen 2.5} & 7B & 8.31 & 8.19 & 0.100 & 0.141\\
         & 14B & 8.29 & 8.18 & 0.003 & 0.114\\
         & 32B & \textbf{8.20} & 8.27 & 0.137 & 0.128\\
         & 72B & 8.31 & 8.16 & 0.064 & 0.112\\
         \midrule
        \multirow{3}{*}{OLMo 2} & 7B & 8.25 & \textbf{8.13} & 0.128 & 0.146\\
         & 13B & 8.26 & 8.21 & \bfseries 0.147 & 0.141\\
         & 32B & 8.27 & 8.27 & 0.142 & 0.101\\
        \midrule
        \multirow{2}{*}{Llama 3.1} & 8B & 8.25 & 8.20 & 0.132 & 0.145\\
         & 70B & 8.24 & 8.16 & 0.134 & \textbf{0.176}\\
         \bottomrule
    \end{tabular}
    \caption{Leave-one-out evaluation results for Low-Lex and High-Lex readers. Selected denotes the internal layer chosen based on $\Delta LL$ on the training texts. MAE values are multiplied by 100 for readability.}
    \label{tab:loto}
\end{table}

\subsection{Illustrative Example}
\setlength{\tabcolsep}{2.5pt}
\begin{table}[t]
    \centering
    \small
    \begin{tabular}{r|cccc|S[table-format=-1.2]cS[table-format=2.2]}
        & \multicolumn{4}{c|}{log-FPGD} & {$r_{G, w}$} & \multicolumn{2}{c}{Surprisal} \\
        Word & Obs. & Surf. & Fin. & Sel. & {Surf.} & Fin. & {Sel.} \\
        \midrule
related & 5.82 & \textbf{5.73} & 5.73 & 5.71 & 0.09 & 1.98 & 2.63 \\
note, & 5.71 & \textbf{5.59} & 5.57 & 5.53 & 0.12 & 0.14 & 0.00 \\
sleep & 5.48 & 5.65 & \textbf{5.63} & 5.66 & -0.17 & 1.32 & 13.01 \\
also & 5.41 & 5.52 & 5.51 & \textbf{5.49} & -0.11 & 1.53 & 0.02 \\
\textit{improves} & 5.63 & 5.88 & \textbf{5.87} & 5.87 & -0.25 & 1.57 & 8.87 \\
concentration & 6.25 & 6.06 & 6.05 & \textbf{6.08} & 0.19 & 2.12 & 7.95 \\
and & 5.34 & 5.37 & \textbf{5.36} & 5.38 & -0.04 & 1.72 & 9.51 \\
mental & 5.75 & \textbf{5.75} & 5.75 & 5.77 & -0.00 & 4.03 & 12.59 \\
alertness. & 6.23 & 6.00 & 5.99 & \textbf{5.97} & 0.23 & 2.08 & 2.48 \\
    \bottomrule
    \end{tabular}
    \caption{Illustrative example for Llama 3.1 70B. The selected internal layer is layer 46 out of 80. \textit{improves} is highlighted as one of the words with the largest residual $r_{G, w}$ under the Surface LMM. Obs. denotes observed log-FPGD, while Surf., Fin., and Sel., denote the Surface, Final, and Selected LMM predictions, respectively. For each word, the prediction closest to the observed value is in bold.}
    \label{tab:loto_ex}
\end{table}

To illustrate the LOO results at the word level, \autoref{tab:loto_ex} shows a example from Llama 3.1 70B, where the layer 46 of 80 is selected with the largest $\Delta LL$ on the training texts. The word \textit{improves} with the largest absolute residual $r_{G, w}$ under the Surface LMM and its surrounding words are shown.

The three LMMs often show similar log-FPGD predictions, which is consistent with the small MAE differences observed in \autoref{tab:loto}. For \textit{improves}, the Surface LMM overestimates its log-FPGD. While surprisal values from the selected and final layers differ substantially, these two LMMs produce similar predictions. A similar pattern can be seen in the surrounding words, where the selected and final layers sometimes assign substantially different surprisal values, while the resulting log-FPGD remain close. This example illustrates that surprisal values can differ across layers, while the resulting differences in predictions remain small.

\section{Discussion}

\subsection{Gaze Measure and Predictive Depth}
One of the primary findings of this study is that Predictive Depth is generally deeper for TGD than for FPGD. This difference is highly consistent across LLMs, occurring in all tested LLMs for High-Lex and in most LLMs for Low-Lex. Since FPGD is restricted to first-pass reading while TGD also includes subsequent fixations, the two measures differ in both the timing and breadth of reading processes they capture. The deeper Predictive Depth observed for TGD suggests that surprisal from deeper layers becomes relatively more predictive as the gaze measure incorporates a broader span of processing.

One possible explanation is that this shift is related to the information encoded at different layer depths. Previous probing studies suggest that deeper LLM layers tend to encode more contextualized information \citep{peters-etal-2018-dissecting, vulic-etal-2020-probing, jin-etal-2025-exploring}. Therefore, TGD may be more sensitive than FPGD to surprisal representations that incorporate broader contextual information, as it includes processing that extends beyond the initial lexical access phase. This interpretation is consistent with the previous findings that relatively fast reading measures tend to be better predicted by shallower layers, whereas slower measures show stronger predictive power in deeper layers \citep{kuribayashi-etal-2025-large,tsipidi-etal-2026-probing}.

Importantly, this pattern does not necessarily imply a direct mapping between layer depth and specific stages of human cognition. Rather, the consistent difference between FPGD and TGD suggests that layer depth may be related to the timing and breadth of the reading processes captured by different gaze measures.

\subsection{Vocabulary Proficiency and Reading Processes}
The difference between High-Lex and Low-Lex readers is more evident in FPGD, where Low-Lex readers tend to show deeper Predictive Depth than High-Lex readers. This difference becomes substantially weaker in TGD. Thus, vocabulary proficiency is more clearly reflected in first-pass reading than in the broader processing captured by TGD.

One possible interpretation is that readers with lower proficiency may rely on contextual information earlier when they encounter linguistic obstacles. Cognitive models of compensatory processing suggest that readers can recruit contextual information when faced with difficult vocabulary or syntax \citep{stanovich-1980-toward}. Given that deeper layers tend to encode more contextualized information, the deeper Predictive Depth observed for Low-Lex readers in FPGD may be consistent with earlier reliance on such information during first-pass reading.

The smaller proficiency difference in TGD is also informative. Since TGD includes later and repeated fixations, it is likely to reflect a later stage of cognitive processing where contextual information has more opportunity to influence gaze behaviour in both proficiency groups. This would reduce the difference between High-Lex and Low-Lex readers in Predictive Depth, even if their first-pass reading patterns differ.

\subsection{Interpreting Predictive Depth}
Predictive Depth should be interpreted as a summary of how the predictive power of surprisal is distributed across layers. A deeper Predictive Depth indicates that predictive power is weighted more towards deeper layers, but it does not imply greater reading difficulty or cognitive load. Likewise, our results do not assume a direct correspondence between LLM layers and specific stages of human cognition.

Instead, the observed findings suggest that layer depth may be informative about differences in the timing and breadth of the reading processes captured by gaze measures. The consistently deeper Predictive Depth for TGD than for FPGD supports this interpretation. Furthermore, the differences in FPGD between two proficiency groups suggest that the layer-wise distribution can vary across reader profiles. Predictive Depth provides a descriptive way to compare these layer-wise patterns without assigning fixed cognitive interpretations to each layer.

\subsection{Transfer Beyond Training Texts}
The leave-one-out analysis provides an external check on whether the layer-wise relationships identified in the main analysis extend beyond the texts used for fitting LMMs. Across LLMs, the selected internal layer generally performs better than the final layer on held-out texts, but the reduction in MAE is quite small in practice. The correlation analysis shows a similar pattern. Residual log-FPGD tends to correlate more strongly with surprisal from the selected internal layer than with that from the final layer, but the correlations remain weak to modest.

These results suggest that the advantage of internal layers is not limited to the texts on which the LMMs were fitted. At the same time, the small MAE improvements and modest correlations indicate that the transferable relationships are limited. The leave-one-out results provide out-of-sample support for the layer-wise analysis, but they do not imply that surprisal alone is sufficient for accurate prediction of gaze behaviour on unseen texts.

\section{Conclusion}

In this paper, we investigate how the layer-wise relationship between surprisal from LLMs and human reading behaviour varies with vocabulary proficiency and gaze measure. With respect to \textbf{RQ1}, vocabulary proficiency is associated with the layer-wise distribution of predictive power, especially for FPGD. Across LLMs, Low-Lex readers tend to show deeper Predictive Depth than High-Lex readers for FPGD, while this difference is substantially smaller for TGD.
Regarding \textbf{RQ2}, the distribution differs systematically between FPGD and TGD. TGD itself generally shows deeper Predictive Depth than FPGD in both groups. These findings suggest that where predictive power is concentrated across LLM layers may be related to the timing and breadth of the reading processes captured by different gaze measures, and that this relationship can vary with reader proficiency.
Our leave-one-out analysis further shows that the advantage of selected internal layers extends to unseen texts, but the overall improvement in prediction is limited.

Future work should examine whether these findings generalize to other languages, datasets, and additional measures of reader proficiency. It will also be important to move beyond group-level comparisons and investigate whether individual reader characteristics can be incorporated directly into models of layer-wise surprisal and gaze behaviour.

\section*{Limitations}
While this study utilizes LexTALE for grouping readers, reading ability is composed of multiple aspects, and vocabulary knowledge is merely one component.
Factors such as background knowledge and working memory capacity also play critical roles in the reading process. In addition, the analysis based on LexTALE groups does not capture continuous variation among readers. Future work could incorporate multiple proficiency measures at the individual level rather than discrete reader groups.

Also, our experiments are based on English reading data from the MECO L2 dataset. The layer-wise relationships observed in this study may be influenced by the specific properties of the English language, particular texts in the corpus, or reader population in the dataset. Languages with different morphological and syntactic properties may show different patterns of layer-wise predictive power. Evaluating the same analysis across languages and datasets will be important for establishing the generalizability of the findings.

The linear mixed-effects models used in this study include random effects to account for repeated observations. However, richer random-effects structures, such as random slopes or word-level predictors, may capture variation that is not represented in our study.

Predictive Depth is a summary statistic calculated from the distribution of $\Delta LL$ across layers. While layer depth is normalized within each LLM, equivalent normalized depth may not represent equivalent information across model architectures. Therefore, comparisons of Predictive Depth across LLMs should be interpreted as comparisons of layer-wise patterns rather than as evidence that particular depths have the same functional role across LLMs.

\section*{Acknowledgments}
This work is partially financed by the Ministerio de Ciencia, Innovación y Universidades, Agencia Estatal de Investigaciones: project CPP2023-010780 funded by MICIU/AEI/10.13039/501100011033 and by FEDER, UE (“Habilitando Modelos de Lenguaje Responsables e Inclusivos”). 
It also received funding from the European Union's Horizon Europe research and innovation program under the Grant Agreement No. 101132431 (iDEM: Innovative and Inclusive Democratic Spaces for Deliberation and Participation). Views and opinions expressed are, however, those of the authors only and do not necessarily reflect those of the European Union. Neither the European Union nor the granting authority can be held responsible for them.

We used ChatGPT for language polishing and proofreading of text originally written by the authors.

\bibliography{custom}

\begin{thebibliography}{34}
\providecommand{\natexlab}[1]{#1}

\bibitem[{Agrawal and Carpuat(2024)}]{agrawal-carpuat-2024-text}
Sweta Agrawal and Marine Carpuat. 2024.
\newblock \href {https://doi.org/10.1162/tacl_a_00653} {Do text simplification systems preserve meaning? a human evaluation via reading comprehension}.
\newblock \emph{Transactions of the Association for Computational Linguistics}, 12:432--448.

\bibitem[{Alva-Manchego et~al.(2020)Alva-Manchego, Scarton, and Specia}]{alva-manchego-etal-2020-data}
Fernando Alva-Manchego, Carolina Scarton, and Lucia Specia. 2020.
\newblock \href {https://doi.org/10.1162/coli_a_00370} {Data-driven sentence simplification: Survey and benchmark}.
\newblock \emph{Computational Linguistics}, 46(1):135--187.

\bibitem[{Baayen et~al.(2008)Baayen, Davidson, and Bates}]{BAAYEN2008390}
R.H. Baayen, D.J. Davidson, and D.M. Bates. 2008.
\newblock \href {https://doi.org/10.1016/j.jml.2007.12.005} {Mixed-effects modeling with crossed random effects for subjects and items}.
\newblock \emph{Journal of Memory and Language}, 59(4):390--412.
\newblock Special Issue: Emerging Data Analysis.

\bibitem[{Carretti et~al.(2009)Carretti, Borella, Cornoldi, and {De Beni}}]{CARRETTI2009246}
Barbara Carretti, Erika Borella, Cesare Cornoldi, and Rossana {De Beni}. 2009.
\newblock \href {https://doi.org/10.1016/j.lindif.2008.10.002} {Role of working memory in explaining the performance of individuals with specific reading comprehension difficulties: A meta-analysis}.
\newblock \emph{Learning and Individual Differences}, 19(2):246--251.

\bibitem[{{Gemma Team}(2025)}]{gemmateam2025gemma3technicalreport}
{Gemma Team}. 2025.
\newblock \href {https://arxiv.org/abs/2503.19786} {Gemma 3 technical report}.
\newblock \emph{Preprint}, arXiv:2503.19786.

\bibitem[{Hale(2001)}]{hale-2001-probabilistic}
John Hale. 2001.
\newblock \href {https://aclanthology.org/N01-1021/} {A probabilistic {E}arley parser as a psycholinguistic model}.
\newblock In \emph{Second Meeting of the North {A}merican Chapter of the Association for Computational Linguistics}.

\bibitem[{Harrington and Sawyer(1992)}]{harrington1992l2}
Michael Harrington and Mark Sawyer. 1992.
\newblock \href {https://doi.org/10.1017/S0272263100010457} {L2 working memory capacity and l2 reading skill}.
\newblock \emph{Studies in second language acquisition}, 14(1):25--38.

\bibitem[{Jin et~al.(2025)Jin, Yu, Huang, Zeng, Wang, Hua, Zhao, Mei, Meng, Ding, Yang, Du, and Zhang}]{jin-etal-2025-exploring}
Mingyu Jin, Qinkai Yu, Jingyuan Huang, Qingcheng Zeng, Zhenting Wang, Wenyue Hua, Haiyan Zhao, Kai Mei, Yanda Meng, Kaize Ding, Fan Yang, Mengnan Du, and Yongfeng Zhang. 2025.
\newblock \href {https://aclanthology.org/2025.coling-main.37/} {Exploring concept depth: How large language models acquire knowledge and concept at different layers?}
\newblock In \emph{Proceedings of the 31st International Conference on Computational Linguistics}, pages 558--573, Abu Dhabi, UAE. Association for Computational Linguistics.

\bibitem[{Kincaid et~al.(1975)Kincaid, Fishburne~Jr, Rogers, and Chissom}]{kincaid1975derivation}
J~Peter Kincaid, Robert~P Fishburne~Jr, Richard~L Rogers, and Brad~S Chissom. 1975.
\newblock Derivation of new readability formulas (automated readability index, fog count and flesch reading ease formula) for navy enlisted personnel.
\newblock \emph{Defense Technical Information Center}.

\bibitem[{Kintsch(1988)}]{kintsch1988role}
Walter Kintsch. 1988.
\newblock \href {https://doi.org/10.1037/0033-295X.95.2.163} {The role of knowledge in discourse comprehension: a construction-integration model}.
\newblock \emph{Psychological review}, 95(2):163--182.

\bibitem[{Kuperman et~al.(2025)Kuperman, Schroeder, Acart{\"u}rk, Agrawal, Alexandre, Bolliger, Brasser, Campos-Rojas, Drieghe, {\DJ}ur{\dj}evi{\'c} et~al.}]{kuperman2025new}
Victor Kuperman, Sascha Schroeder, Cengiz Acart{\"u}rk, Niket Agrawal, Dominick~M Alexandre, Lena~S Bolliger, Jan Brasser, C{\'e}sar Campos-Rojas, Denis Drieghe, Du{\v{s}}ica~Filipovi{\'c} {\DJ}ur{\dj}evi{\'c}, and 1 others. 2025.
\newblock \href {https://doi.org/10.1017/S0272263125000105} {New data on text reading in english as a second language: The wave 2 expansion of the multilingual eye-movement corpus (meco)}.
\newblock \emph{Studies in Second Language Acquisition}, 47(2):677--695.

\bibitem[{Kuperman et~al.(2023)Kuperman, Siegelman, Schroeder, Acart{\"u}rk, Alexeeva, Amenta, Bertram, Bonandrini, Brysbaert, Chernova et~al.}]{kuperman2023text}
Victor Kuperman, Noam Siegelman, Sascha Schroeder, Cengiz Acart{\"u}rk, Svetlana Alexeeva, Simona Amenta, Raymond Bertram, Rolando Bonandrini, Marc Brysbaert, Daria Chernova, and 1 others. 2023.
\newblock \href {https://doi.org/10.1017/S0272263121000954} {Text reading in english as a second language: Evidence from the multilingual eye-movements corpus}.
\newblock \emph{Studies in second language acquisition}, 45(1):3--37.

\bibitem[{Kuribayashi et~al.(2022)Kuribayashi, Oseki, Brassard, and Inui}]{kuribayashi-etal-2022-context}
Tatsuki Kuribayashi, Yohei Oseki, Ana Brassard, and Kentaro Inui. 2022.
\newblock \href {https://doi.org/10.18653/v1/2022.emnlp-main.712} {Context limitations make neural language models more human-like}.
\newblock In \emph{Proceedings of the 2022 Conference on Empirical Methods in Natural Language Processing}, pages 10421--10436, Abu Dhabi, United Arab Emirates. Association for Computational Linguistics.

\bibitem[{Kuribayashi et~al.(2025)Kuribayashi, Oseki, Taieb, Inui, and Baldwin}]{kuribayashi-etal-2025-large}
Tatsuki Kuribayashi, Yohei Oseki, Souhaib~Ben Taieb, Kentaro Inui, and Timothy Baldwin. 2025.
\newblock \href {https://doi.org/10.1162/TACL.a.58} {Large language models are human-like internally}.
\newblock \emph{Transactions of the Association for Computational Linguistics}, 13:1743--1766.

\bibitem[{Kuribayashi et~al.(2026)Kuribayashi, Warstadt, Oseki, and Wilcox}]{kuribayashi-etal-2026-dual}
Tatsuki Kuribayashi, Alex Warstadt, Yohei Oseki, and Ethan~Gotlieb Wilcox. 2026.
\newblock \href {https://doi.org/10.18653/v1/2026.acl-long.2143} {Dual alignment between language model layers and human sentence processing}.
\newblock In \emph{Proceedings of the 64th Annual Meeting of the {A}ssociation for {C}omputational {L}inguistics (Volume 1: Long Papers)}, pages 46207--46223, San Diego, California, United States. Association for Computational Linguistics.

\bibitem[{LaBerge and Samuels(1974)}]{laberge1974toward}
David LaBerge and S~Jay Samuels. 1974.
\newblock \href {https://doi.org/10.1016/0010-0285(74)90015-2} {Toward a theory of automatic information processing in reading}.
\newblock \emph{Cognitive psychology}, 6(2):293--323.

\bibitem[{Lemh{\"o}fer and Broersma(2012)}]{lemhofer2012introducing}
Kristin Lemh{\"o}fer and Mirjam Broersma. 2012.
\newblock \href {https://doi.org/10.3758/s13428-011-0146-0} {Introducing lextale: A quick and valid lexical test for advanced learners of english}.
\newblock \emph{Behavior research methods}, 44(2):325--343.

\bibitem[{Levy(2008)}]{levy2008expectation}
Roger Levy. 2008.
\newblock \href {https://doi.org/10.1016/j.cognition.2007.05.006} {Expectation-based syntactic comprehension}.
\newblock \emph{Cognition}, 106(3):1126--1177.

\bibitem[{{Llama Team}(2024)}]{grattafiori2024llama3herdmodels}
{Llama Team}. 2024.
\newblock \href {https://arxiv.org/abs/2407.21783} {The llama 3 herd of models}.
\newblock \emph{Preprint}, arXiv:2407.21783.

\bibitem[{Oh et~al.(2022)Oh, Clark, and Schuler}]{oh2022comparison}
Byung-Doh Oh, Christian Clark, and William Schuler. 2022.
\newblock \href {https://doi.org/10.3389/frai.2022.777963} {Comparison of structural parsers and neural language models as surprisal estimators}.
\newblock \emph{Frontiers in Artificial Intelligence}, 5:777963.

\bibitem[{Oh and Schuler(2023)}]{oh-schuler-2023-surprisal}
Byung-Doh Oh and William Schuler. 2023.
\newblock \href {https://doi.org/10.1162/tacl_a_00548} {Why does surprisal from larger transformer-based language models provide a poorer fit to human reading times?}
\newblock \emph{Transactions of the Association for Computational Linguistics}, 11:336--350.

\bibitem[{{OLMo Team}(2025)}]{olmo20252olmo2furious}
{OLMo Team}. 2025.
\newblock \href {https://arxiv.org/abs/2501.00656} {2 olmo 2 furious}.
\newblock \emph{Preprint}, arXiv:2501.00656.

\bibitem[{Perfetti and Hart(2002)}]{perfetti2002lexical}
Charles~A Perfetti and Lesley Hart. 2002.
\newblock \href {https://doi.org/10.1075/swll.11.14per} {The lexical quality hypothesis}.
\newblock \emph{Precursors of functional literacy}, 11:67--86.

\bibitem[{Peters et~al.(2018)Peters, Neumann, Zettlemoyer, and Yih}]{peters-etal-2018-dissecting}
Matthew~E. Peters, Mark Neumann, Luke Zettlemoyer, and Wen-tau Yih. 2018.
\newblock \href {https://doi.org/10.18653/v1/D18-1179} {Dissecting contextual word embeddings: Architecture and representation}.
\newblock In \emph{Proceedings of the 2018 Conference on Empirical Methods in Natural Language Processing}, pages 1499--1509, Brussels, Belgium. Association for Computational Linguistics.

\bibitem[{{Qwen Team}(2025)}]{qwen2025qwen25technicalreport}
{Qwen Team}. 2025.
\newblock \href {https://arxiv.org/abs/2412.15115} {Qwen2.5 technical report}.
\newblock \emph{Preprint}, arXiv:2412.15115.

\bibitem[{Rayner(1998)}]{rayner1998eye}
Keith Rayner. 1998.
\newblock \href {https://doi.org/10.1037/0033-2909.124.3.372} {Eye movements in reading and information processing: 20 years of research.}
\newblock \emph{Psychological bulletin}, 124(3):372.

\bibitem[{Seabold and Perktold(2010)}]{seabold2010statsmodels}
Skipper Seabold and Josef Perktold. 2010.
\newblock \href {https://doi.org/10.25080/majora-92bf1922-011} {statsmodels: Econometric and statistical modeling with python}.
\newblock In \emph{9th Python in Science Conference}.

\bibitem[{Shain et~al.(2024)Shain, Meister, Pimentel, Cotterell, and Levy}]{shain2024large}
Cory Shain, Clara Meister, Tiago Pimentel, Ryan Cotterell, and Roger Levy. 2024.
\newblock \href {https://doi.org/10.1073/pnas.2307876121} {Large-scale evidence for logarithmic effects of word predictability on reading time}.
\newblock \emph{Proceedings of the National Academy of Sciences}, 121(10):e2307876121.

\bibitem[{Siddharthan(2014)}]{siddharthan2014survey}
Advaith Siddharthan. 2014.
\newblock \href {https://doi.org/10.1075/itl.165.2.06sid} {A survey of research on text simplification}.
\newblock \emph{ITL-International Journal of Applied Linguistics}, 165(2):259--298.

\bibitem[{Smith and Levy(2013)}]{smith2013effect}
Nathaniel~J Smith and Roger Levy. 2013.
\newblock \href {https://doi.org/10.1016/j.cognition.2013.02.013} {The effect of word predictability on reading time is logarithmic}.
\newblock \emph{Cognition}, 128(3):302--319.

\bibitem[{Speer(2022)}]{robyn_speer_2022_7199437}
Robyn Speer. 2022.
\newblock \href {https://doi.org/10.5281/zenodo.7199437} {rspeer/wordfreq: v3.0}.

\bibitem[{Stanovich(1980)}]{stanovich-1980-toward}
Keith~E. Stanovich. 1980.
\newblock \href {http://www.jstor.org/stable/747348} {Toward an interactive-compensatory model of individual differences in the development of reading fluency}.
\newblock \emph{Reading Research Quarterly}, 16(1):32--71.

\bibitem[{Tsipidi et~al.(2026)Tsipidi, Kiegeland, Re, Xu, Giulianelli, Stanczak, and Cotterell}]{tsipidi-etal-2026-probing}
Eleftheria Tsipidi, Samuel Kiegeland, Francesco~Ignazio Re, Tianyang Xu, Mario Giulianelli, Karolina Stanczak, and Ryan Cotterell. 2026.
\newblock \href {https://doi.org/10.18653/v1/2026.acl-long.575} {Probing for reading times}.
\newblock In \emph{Proceedings of the 64th Annual Meeting of the {A}ssociation for {C}omputational {L}inguistics (Volume 1: Long Papers)}, pages 12618--12642, San Diego, California, United States. Association for Computational Linguistics.

\bibitem[{Vuli{\'c} et~al.(2020)Vuli{\'c}, Ponti, Litschko, Glava{\v{s}}, and Korhonen}]{vulic-etal-2020-probing}
Ivan Vuli{\'c}, Edoardo~Maria Ponti, Robert Litschko, Goran Glava{\v{s}}, and Anna Korhonen. 2020.
\newblock \href {https://doi.org/10.18653/v1/2020.emnlp-main.586} {Probing pretrained language models for lexical semantics}.
\newblock In \emph{Proceedings of the 2020 Conference on Empirical Methods in Natural Language Processing (EMNLP)}, pages 7222--7240, Online. Association for Computational Linguistics.

\end{thebibliography}

\clearpage
\appendix
\onecolumn
\section{Plots of Predictive Power of Layer-wise Surprisal}
\label{sec:plot_surprisal}

\begin{figure*}[ht]
    \centering
    \includegraphics[width=0.95\linewidth]{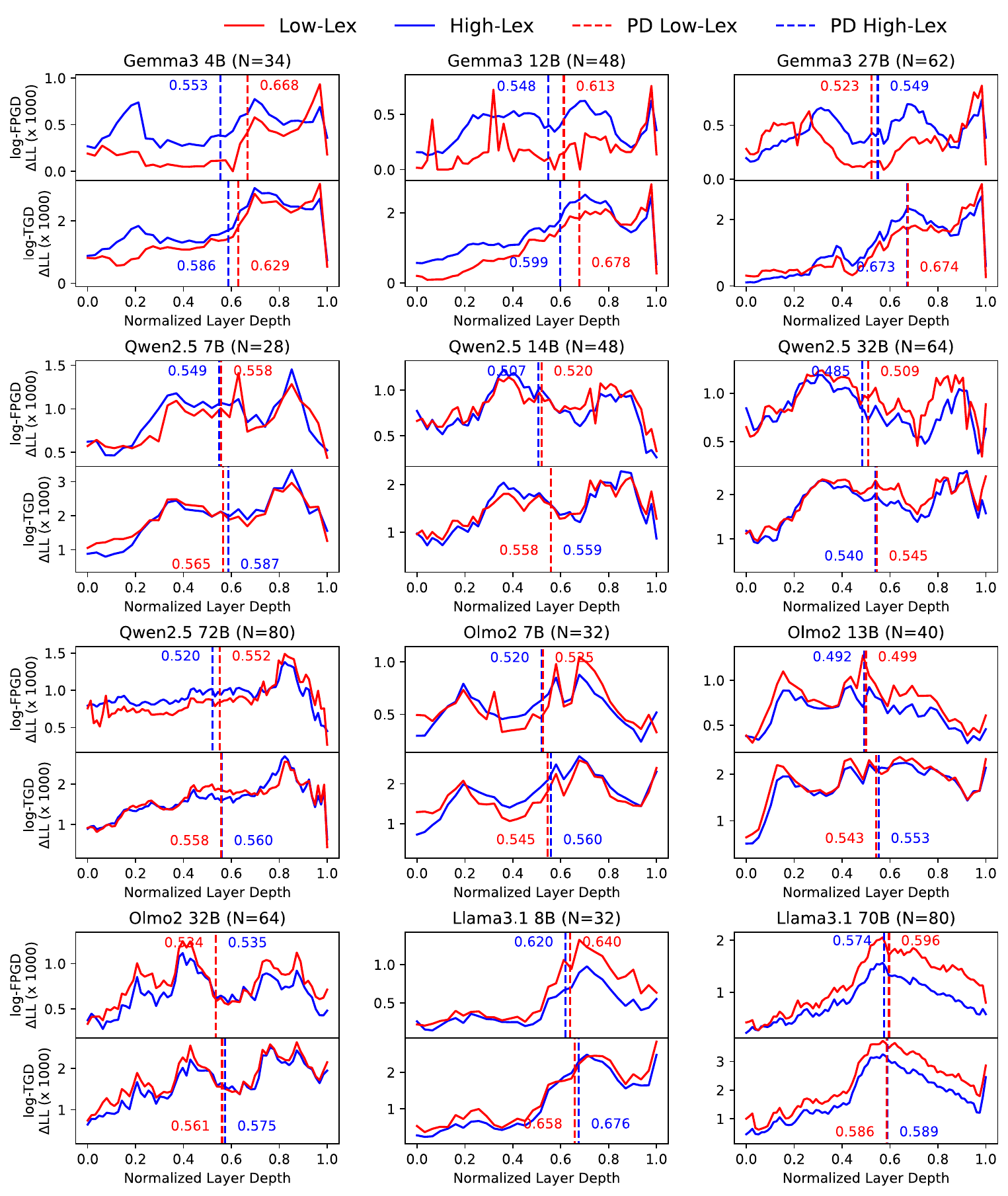}
    \caption{$\Delta LL$ and Predictive Depth across all tested LLMs. As in \autoref{fig:deltall}, the color blue refers to High-Lex, and the red refers to Low-Lex.}
    \label{fig:delta_ll_appendix}
\end{figure*}

\twocolumn
\section{Implementation Details}

\subsection{Surprisal Extraction}
We use HuggingFace Transformers library to extract the log-probabilities for calculating surprisal values. All inputs are processed as plain text without additional prompt templates. While the interest areas of eye-tracking measurements often include adjacent punctuation marks, such as periods or quotation marks, we exclude these symbols from our surprisal analysis to focus on linguistic and cognitive processing of lexical items. In other words, for words followed by punctuation, its surprisal is calculated based solely on the constituent characters of the word itself.

\subsection{LMM Settings}
To fit the linear mixed-effects models, we use the default optimizer \textit{lbfgs} (Limited-memory BFGS) in the \texttt{statsmodels} library in Python. The maximum number of iterations is set to 1000. Other hyperparameters are set to default values.

\end{document}